%% file: eacl2023.tex
\pdfoutput=1

\documentclass[11pt]{article}

\usepackage[]{EACL2023}

\usepackage{svg}

\usepackage{enumitem}

\usepackage{times,multirow}
\usepackage{latexsym,graphicx,amsmath,subfigure}

\usepackage[T1]{fontenc}

\usepackage[utf8]{inputenc}

\usepackage{microtype}

\newcommand{\m}{{\textsc{PromptDA}}}

\usepackage{microtype}

\usepackage{array}
\usepackage{booktabs}

\usepackage{capt-of}
\usepackage{colortbl}

\title{{\m} : Label-guided Data Augmentation \\ for Prompt-based Few-shot Learners}

\author{Canyu Chen \\
  Illinois Institute of Technology \\
  \texttt{cchen151@hawk.iit.edu} \\\And
  Kai Shu \\
  Illinois Institute of Technology \\
  \texttt{kshu@iit.edu} \\}

\begin{document}
\maketitle
\begin{abstract}

Recent advances in large pre-trained language models (PLMs) lead to impressive gains in natural language understanding (NLU) tasks with task-specific fine-tuning. However, directly fine-tuning PLMs heavily relies on sufficient labeled training instances, which are usually hard to obtain. Prompt-based tuning on PLMs has shown to be powerful for various downstream few-shot tasks. Existing works studying prompt-based tuning for few-shot NLU tasks mainly focus on deriving proper label words with a verbalizer or generating prompt templates to elicit semantics from PLMs. 
In addition, conventional data augmentation strategies such as synonym substitution, though widely adopted in low-resource scenarios, only bring marginal improvements for prompt-based few-shot learning.
Thus, an important research question arises: \textit{how to design effective data augmentation methods for prompt-based few-shot tuning?} To this end, considering the label semantics are essential in prompt-based tuning, we propose a novel \emph{label-guided data augmentation} framework \textbf{\textsc{PromptDA}}, which exploits the enriched label semantic information for data augmentation. Extensive experiment results on few-shot text classification tasks 
demonstrate the superior performance of the proposed framework
by effectively leveraging label semantics and data augmentation for natural language understanding. Our code is available at \href{https://github.com/canyuchen/PromptDA}{https://github.com/canyuchen/PromptDA}.

\end{abstract}

\input{intro_v2}

\input{related}
\input{framework}

\input{experiments}

\section{Conclusion and Future Work}
In this paper, we study a new problem of \textit{data augmentation in prompt-based tuning for few-shot learners}. To leverage the label semantic information, we propose a novel label-guided data augmentation approach {\m}, which can derive multiple label words for each class and naturally incorporate them into data augmentation for prompt tuning. We conduct extensive experiments on various  datasets and demonstrate the effectiveness of {\m} in few-shot scenarios. 
We also conduct detailed analysis on the effects of  manual or automatic label augmentation, the size of augmentation, the size of the training set, comparison, and combination with conventional DA.

There are multiple interesting directions for future work. First, we will further investigate how to fully exploit the semantic label space for enhancing the \textit{performance and stability} of prompt-based tuning. Second, we will extend {\m} to multi-label few-shot tasks and leverage multi-aspect label space. Third, we will explore data augmentation strategies for prompt-based tuning in more applications such as information extraction, intent detection, text matching, language generation, and social computing.

\section*{Limitations}

Our work is the first step in designing data augmentation strategies for the prompt tuning paradigm. In this work, we only focus on the natural language understanding (NLU) tasks. The prompt tuning paradigm is also widely applied in various tasks including language generation, question answering, dialog systems, etc. 
Designing augmentation strategies for prompt-based few-shot learners in more applications is under exploration. Our work can inspire future research on utilizing label semantics for improving prompt tuning.

\section*{Ethics Statement}

This paper focuses on the task of few-shot natural language understanding and conducts experiments on open datasets.
The implementation details are described in Appendix for reproduction. Because the prompt-based tuning methodology only utilizes very few training instances,  it may inherit undesired biases that stem from language model pretraining on a large corpus.

\section*{Acknowledgements}
We would like to appreciate the valuable feedback from anonymous reviewers. This work is supported in part by NSF SaTC-2241068, and a Cisco Research Award.

\newpage 
\clearpage

\bibliographystyle{acl_natbib}
\bibliography{main_2}

\appendix
\input{appendix}

\end{document}

%% file: intro_v2.tex
\section{Introduction}

\begin{figure}[!tbp]
    \centering
    \includegraphics[width=.47\textwidth]{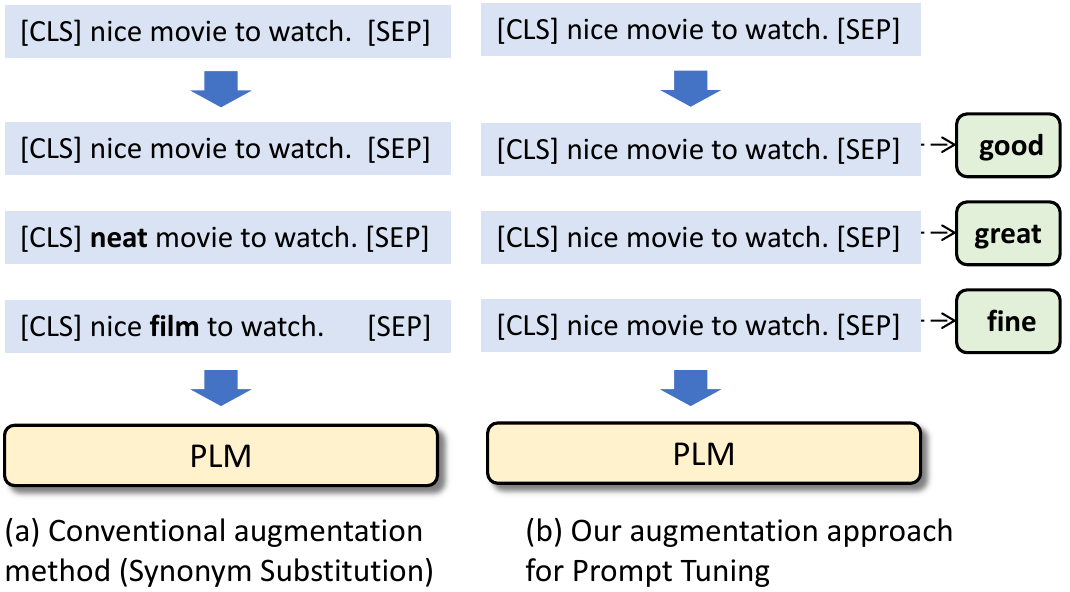}
    \caption{The basic comparison of conventional data augmentation methods and our proposed augmentation framework \textbf{{\m}}. $\{\texttt{good, great, fine}\}$ are the label words for prompt tuning. Conventional DA constructs \textbf{\textit{instances}} for augmentation, but {\m} constructs \textbf{\textit{instance-label pairs}} instead.}
    \label{fig:intro_figure}
    \vspace{-0.5cm}
\end{figure}

Pre-trained language models (PLMs) have shown promising performance in various applications such as text classification~\cite{yang2019xlnet}, document summarization~\cite{zhang2020pegasus}, question answering~\cite{mirzaee2021spartqa}. 
The recent advancement of \emph{prompt-based tuning} has shown a significant improvement over normal fine-tuning on various few-shot tasks~\cite{brown2020language}.
Typically, the prompt-based tuning paradigm transforms an NLU task into a masked language modeling (MLM) problem. For example, in sentiment analysis, an original sentence ``\texttt{nice movie to watch.}" can be augmented with a \emph{template} ``\texttt{It is [MASK]}" as the input $x$. Each class (e.g., \textsc{Positive}) is represented by a \emph{label word} (e.g., \texttt{good}) selected by a \emph{verbalizer} from the vocabulary~\cite{schick2021exploiting}. 
The prediction of the class \textsc{Positive} is based on the probability of the {\verb|[MASK]|} being filled with the token \texttt{good}.

In addition, conventional data augmentation (DA) methods such as synonym substitution are also widely applied when the training data is limited~\cite{chen2021empirical}. However, it has been shown that they can only bring marginal improvements for prompt-based few-shot learning~\cite{zhou2021flipda}.
We argue that one of the reasons could be that these data augmentation methods mainly focus on transforming the \emph{instances} while ignoring the \emph{label semantics}, which have great potential to improve the performance of few-shot tasks~\cite{luo2021don} and are essential for prompt-based few-shot learners~\cite{liu2021pre}. 
Therefore, we focus on a new problem of designing augmentation strategies for the prompt tuning paradigm and explore fusing label semantics into data augmentation for prompt-based few-shot learners.

Specifically, different from most prompt-based tuning methods that adopt a \emph{one-to-one verbalizer}~\cite{schick2021exploiting, gao2021making}, we propose to incorporate the rich label semantic information derived from a  \emph{one-to-multiple verbalizer} into a new data augmentation paradigm. As shown in Figure~\ref{fig:intro_figure}, compared with previous data augmentation approaches that mainly focus on constructing more \textbf{\emph{instances}}, {\m}  proposes to construct \textbf{\emph{instance-label pairs}} for augmentation, which opens a new dimension for conducting augmentation. 
For example, with a one-to-multiple verbalizer mapping from the class \textsc{Positive} to a set of  label words $\{\texttt{good, great, fine}\}$, we aim to generate a set of synthetic data points $ \{(x, \texttt{good}), (x, \texttt{great}), (x, \texttt{fine}) \}$ based on  the original instance $x$ and the label words, and then leverage them to enhance the performance of the  prompt-based few-shot learners.
Furthermore, extensive experiment results in Section~\ref{sec:Combination_with_Conventional_DA} show that {\m} can be regarded \textit{orthogonal} to the conventional DA methods (e.g., synonym substitution). Thus, {\m} can complement conventional augmentation approaches to further improve the performance.

In essence, we propose a new  \textit{label-guided data augmentation framework for  prompt-based few-shot learners} named {\m},  which contains three coherent modules including \textit{Label Augmentation}, \textit{Augmented Prompt-based Tuning}, and \textit{Prediction Transformation}. First, we utilize a PLM to automatically search for a one-to-multiple verbalizer on a specific training set, and then use it to derive a set of semantically similar tokens for each class as the corresponding label words. Second, in the training stage, we construct a set of instance-label pairs from the original data with regard to each label word for augmentation in prompt tuning. 
Third, in the inference stage, we utilize the trained language model to predict the label by aggregating the probability scores on the derived label words.

The contributions of this paper are summarized as follows:
(1) we study a new problem of designing data augmentation strategies for prompt-based few-shot learners;
(2) we propose a novel label-guided data augmentation framework {\m} that exploits the rich label semantic information of a one-to-multiple verbalizer for improving prompt tuning;
(3) we conduct extensive experiments on real-world few-shot classification tasks and demonstrate the effectiveness of the proposed framework.

%% file: related.tex
\section{Related Work}

\paragraph{Prompt-based Tuning} has attracted increasing attention recently for various natural language processing tasks including natural language understanding~\cite{gao2021making,Liu2022PTuningPT}, question answering~\cite{jiang2020can,wang-etal-2022-kecp}, language generation~\cite{li2021prefix,li-etal-2022-learning-transfer}, information extraction~\cite{DBLP:conf/www/ChenZXDYTHSC22,son2022grasp}, machine translation~\cite{tan2021msp}, dialog systems~\cite{zhu-etal-2022-continual,huang2022prompt}, social computing~\cite{HUANG2023103279,lin2022zeroshot}, etc. The prompt-based learning framework has shown promising performance, especially in few-shot or zero-shot classification tasks when limited or no labeled data is available~\cite{liu2021pre}. For example, Gao \textit{et al.} propose a prompt-based fine-tuning framework that automatically generates prompt templates and demonstrations to improve few-shot classification performance~\cite{gao2021making}. Shin \textit{et al.} propose AutoPrompt to automatically generate prompts and verbalizers for eliciting the knowledge from language models~\cite{shin2020autoprompt}. Other works on improving prompt-based model performance also mainly focus on constructing various types of prompt templates and verbalizers~\cite{liu2021pre}.

\paragraph{Few-shot Text Classification} aims to build a text classification model when few labeled instances are available. Generally, the few-shot text classification problem is connected to the following learning paradigms: (1) semi-supervised learning where unlabeled data, alongside a small amount of labeled data, is used for learning~\cite{mukherjee2020uncertainty,lee2021salnet}. For example, Subhabrata \textit{et al.} propose to jointly learn from limited labeled data and a large amount of unlabeled data with uncertainty-aware self-training ~\cite{mukherjee2020uncertainty}. (2) meta-learning frameworks such as metric-based~\cite{sui2020knowledge} and optimization-based approaches~\cite{bansal2019learning,yin2020meta}. (3) weakly supervised learning where weak labels are derived~\cite{shu2020learning,meng2020text} in addition to the limited clean labels to improve text classification.
Other few-shot text classification approaches include transfer learning from the source domain to the target domain~\cite{gupta2020effective}, or leveraging auxiliary tasks to improve the target tasks~\cite{xia2021metaxl}.

\paragraph{Data Augmentation} is to construct synthetic samples from an available dataset to enlarge the data size, which can help supervised training~\cite{ren-etal-2021-text,shu2018deep} or self-supervised training~\cite{DBLP:conf/ijcai/BhattacharjeeK022}.  Data augmentation techniques for natural language generally fall into \textit{data space} or \textit{feature space}~\cite{bayer2021survey}.  In the \textit{data space}, augmentation methods transform the instances in  character-level, word-level, phrase-level or document-level~\cite{wei-zou-2019-eda,zhang2022treemix,zhao-etal-2022-epida,kong-etal-2022-dropmix,yang-etal-2020-planning,yang-etal-2020-generative,liu-etal-2020-data,chen-liu-2022-rethinking}. In the \textit{feature space}, representations in the latent space are manipulated by adding noise or interpolation~\cite{guo2019augmenting,sawhney-etal-2022-ciaug,chen2020mixtext,zhang-etal-2020-seqmix,cao2022survey,shen2020simple,si-etal-2021-better,wu-etal-2022-text,chen-etal-2022-doublemix}.

However, conventional augmentation methods can only bring marginal improvements under the prompt tuning paradigm~\cite{zhou2021flipda}.
It is under exploring about how to design effective augmentation methods for prompt-based few-shot scenarios. Therefore, in this paper, we propose a novel \textit{label-guided data augmentation mechanism for prompt-based tuning in few-shot tasks}.
It is worth noting that our proposed {\m} is different from the previous work~\cite{DBLP:conf/acl/0003XSHTGJ22} which utilizes prompt tuning to generate instances for NLU models. {\m} is a new augmentation paradigm for prompt-based learners and can be a \textit{plug-in} module for any prompt tuning method.

%% file: framework.tex
\section{Problem Definition}
The goal of the few-shot classification task is to learn a classifier to predict the label of unseen instances with limited labeled samples during the training. 
Following the widely-used few-shot setting~\cite{gao2021making,liu2021pre}, we assume that a pre-trained language model (e.g., BERT) $\mathcal{M}$ can be utilized to fine-tune on a downstream task with the dataset $\mathcal{D}=\{\mathcal{X},\mathcal{Y}\}$, where $\mathcal{X}$ denotes the instances and $\mathcal{Y}$ indicates the corresponding labels. For each task, the number of training instances for each class is $K$, which is usually small (e.g., 8). The goal is to design a prompt learning strategy that generalizes well on the test set $\mathcal{D}_{\text{test}}$ with few labeled training data in $\mathcal{D}_{\text{train}}$.
To ensure a fair performance evaluation, we assume that a validation set $\mathcal{D}_{val}$ is available, and $|\mathcal{D}_{\text{\text{val}}}|=|\mathcal{D}_{\text{train}}|$. The test set $\mathcal{D}_{\text{test}}$ is the same as the full-data training setting.

\section{Label-guided Data Augmentation for Prompt-based Tuning}
\begin{figure*}[tbp!]
   \centering
   \includegraphics[width=1\textwidth]{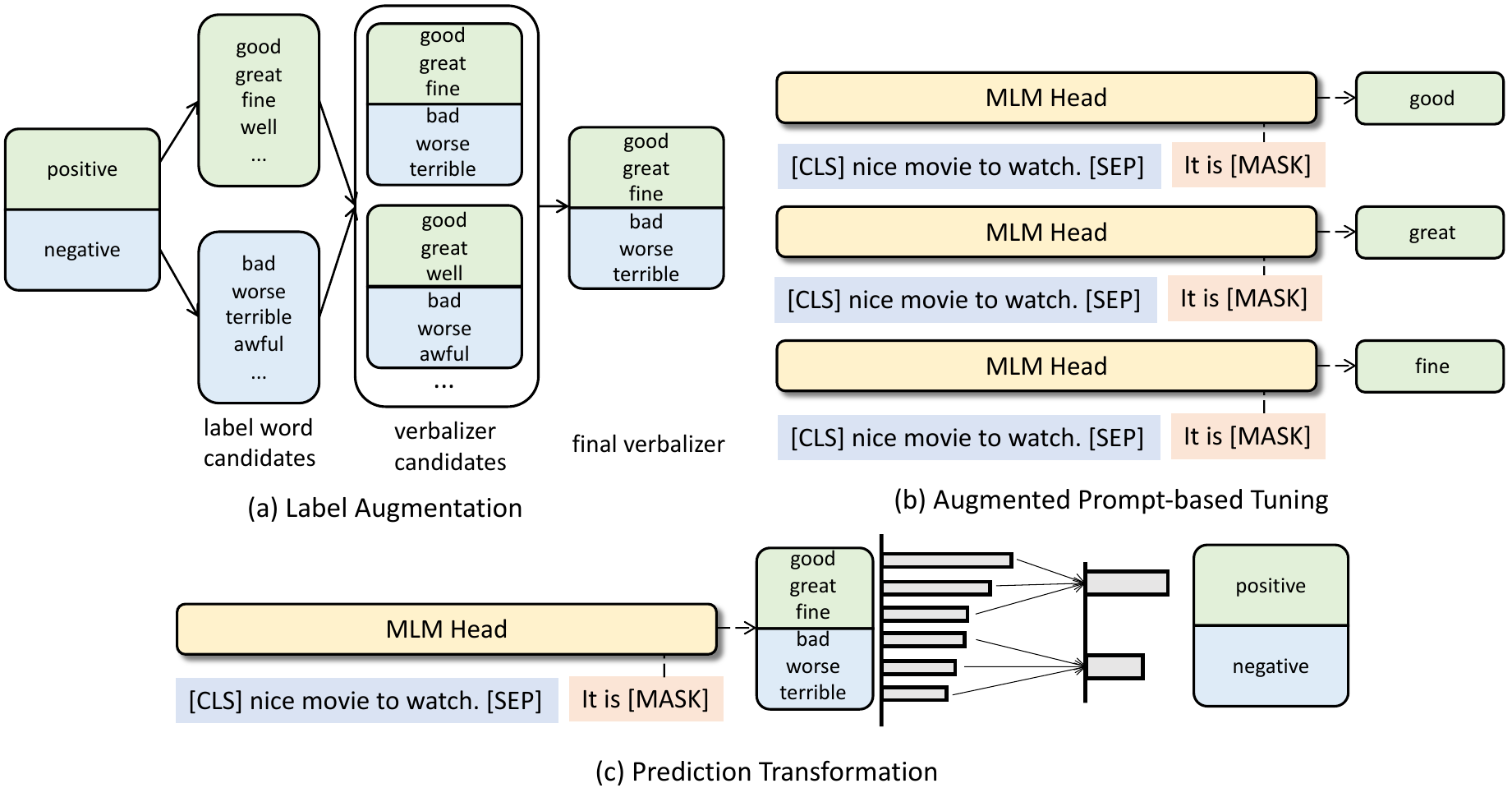}
   \caption{The proposed \textbf{{\m}} for few-shot learning (with sentiment classification task as an example): (a) \textit{Label Augmentation}: deriving multiple label words for each class to enrich the label semantic space; (b) \textit{Augmented Prompt-based Tuning}: training with the augmented instance-label pairs via masked language modeling; (c) \textit{Prediction Transformation}:  aggregating  the probability scores on the derived label words for the final prediction. }
   \label{fig:promptda}
\end{figure*}

In this section, we detail the proposed framework  {\m}, which is illustrated in Figure~\ref{fig:promptda}. It mainly consists of three modules: (1) a \textit{Label Augmentation} module to derive multiple label words for each class to enrich the label space; (2) an \textit{Augmented Prompt-based Tuning} module for augmenting the training data guided by label words; (3) a \textit{Prediction Transformation} module to transform the prediction from the label words to original classes.

\subsection{Label Augmentation}

Due to the limited available labels in few-shot learning, recent works are generating \textit{label words} to help prediction~\cite{schick2021exploiting,gao2021making}. The goal is to extend the label space by incorporating the rich semantics in the vocabulary. The existing works mainly focus on selecting one label word for each class manually or automatically in prompt tuning, where the resultant label words often have a large variance and the semantics in other candidate label words are ignored. Therefore, we explore automatically searching for multiple label words for each class to better enrich the label space. 
Let $\mathcal{F}: \mathcal{Y} \rightarrow  \mathcal{V}_\mathcal{Y}$ denote the \textit{one-to-multiple verbalizer} that maps each label category $y \in \mathcal{Y} $ to  a set of label words in the vocabulary $ \mathcal{V}_y = \{v_y^1, v_y^2, ..., v_y^{k_y}\} \subset \mathcal{V}_\mathcal{Y} $,  where $k_y = | \mathcal{V}_y|$ denotes the number of selected label words for each class. Generally, our proposed search process has two steps including \textit{searching for label word candidates} and \textit{searching for verbalizer candidates}.

Firstly, we aim to search for a set of label word candidates $\tilde{\mathcal{V}}_y \subset \mathcal{V}_\mathcal{Y} $ that are semantically similar to each class  $y\in\mathcal{Y}$.  Let $\mathcal{D}^y_{\text{train}}$ denote the subset of training data with the class $y$. $\mathcal{T}(x)$ denotes the input $x$ with a fixed template $\mathcal{T}$. $\texttt{Po([mask])}$ denotes the position of $\texttt{[mask]}$ in the input $x$. 
We propose to select the $\operatorname{Top-}m$ label words from the vocabulary as $\tilde{\mathcal{V}}_y$ based on the conditional likelihood over $\mathcal{D}^{y}_{\text{train}}$ for each class $y$:
\begin{equation}
\tilde{\mathcal{V}}_y
=
\underset{v \in \mathcal{V}}{\operatorname{Top-}m}
\left
\{\sum_{(x,y) \in \mathcal{D}^{y}_{\text{train}}}\mathrm{Pr}(v, \mathcal{T}(x))
\right\}
\end{equation}

where $\mathrm{Pr}(v, \mathcal{T}(x))$ denotes the corresponding probability score of each token in the vocabulary filling in $\texttt{Po([mask])}$ in PLM inference as:
\begin{equation}
\mathrm{Pr}(v, \mathcal{T}(x)) = \mathrm{Pr}(\texttt{Po([mask])} = v \mid \mathcal{T}(x))
\end{equation}

Secondly, we construct a verbalizer candidate set $F$ for the whole dataset. It is a combinatorial problem to select $k_y$ label words from $\tilde{\mathcal{V}}_y$ to construct $\mathcal{V}_y$ for each class $y$. The number of candidates of  $\mathcal{V}_y$  is $\binom{|\tilde{\mathcal{V}}_y|}{k_y}$. Then the element number of the verbalizer candidate set $F$ is $ |F| = \binom{|\tilde{\mathcal{V}}_y|}{k_y}^{|\mathcal{Y}|}$. We utilize each one-to-multiple verbalizer candidate in $F$ to infer on  $\mathcal{D}_{\text{train}}$ via the same \emph{prediction transformation} method in Section \ref{sec:Prediction_Transformation} and calculate the prediction accuracy. Then we select the ${\operatorname{Top-}n}$ candidates from $F$ based on the prediction accuracy.  If there exist multiple candidates with the same accuracy score, we randomly select one as the final one-to-multiple verbalizer. Otherwise, we select the verbalizer candidate with the highest accuracy score. Note that $m$ and $n$ are both hyperparameters picked by a pilot study on the specific training sets.

\subsection{Augmented Prompt-based Tuning}
To enrich the training data for the few-shot text classification task, it is natural to utilize data augmentation methods such as token-level or sentence-level augmentation for fine-tuning~\cite{DBLP:conf/acl/FengGWCVMH21,li2021data,pluscec2023data}.  Most of the existing data augmentation methods  focus on enlarging training data conditioned on the original label space. 
Orthogonal to previous augmentation methods, our method incorporates label semantic information into prompt-tuning via augmenting instance-label pairs rather than only augmenting instances.
For $(x, y) \in \mathcal{D}_{\text{train}}$, we have obtained the corresponding label word set  $\mathcal{V}_y = \{v_y^1, v_y^2, ..., v_y^{k_y}\}$.
Then we can include $ \{(x,v_y^1), (x,v_y^2), ..., (x,v_y^{k_y})\}$ for augmentation. 
Let $\tilde{\mathcal{D}}_{\text{train}}$ denote the augmented dataset.
The resultant dataset can be denoted as:
\begin{equation}
\tilde{\mathcal{D}}_{\text{train}} = \cup_{(x,y) \in \mathcal{D}_{\text{train}}} \{(x,v_y^1), (x,v_y^2), ..., (x,v_y^{k_y})\}
\end{equation}
In the training process, we follow the MLM training paradigm and minimize the negative log-likelihood on the whole augmented training set $\tilde{\mathcal{D}}_{\text{train}}$.
The optimization objective is:
\begin{equation}
 \mathcal{L} = \sum_{(x, v) \in \tilde{\mathcal{D}}_{\text{train}}} - \log \mathrm{Pr}(v \mid x )
\end{equation}
For $(x, v) \in \tilde{\mathcal{D}}_{\text{train}}$,  the conditional probability of filling the position of $\texttt{[mask]}$ with $v$ is:
\begin{equation}
\begin{aligned}
\mathrm{Pr}(v \mid x ) 
&=\mathrm{Pr}(\texttt{Po([mask])} = v \mid x ) \\
&=\frac{\exp \left(\mathbf{w}_{v} \cdot \mathbf{h}_{\texttt {[MASK]}}\right)}
{\sum_{v' \in \mathcal{V}} \exp \left(\mathbf{w}_{v'} \cdot \mathbf{h}_{\texttt {[MASK]}}\right)}
\end{aligned}
\end{equation}
where $\mathbf{w}_{v}$ denotes the pre-softmax output vector for each token $v$ in the vocabulary, and $\mathbf{h}_{\texttt {[MASK] }}$ denotes the corresponding hidden state of the $\texttt{[MASK]}$  position. Note that we completely reuse the PLM and do not introduce new parameters in the training process, which is important for prompt-based tuning  to be effective in few-shot scenarios.

\subsection{Prediction Transformation}
\label{sec:Prediction_Transformation}

We have introduced the process of training an MLM classifier head with the augmented data in prompt-based tuning. Next, we describe how to perform inference for the target class.
Let $h$ denote the function that transforms the probability scores on the label word set  $\mathcal{V}_y = \{v_y^1, v_y^2, ..., v_y^{k_y}\}$ into the probability score of each class $y$. Since the label word with the highest probability score in set $\mathcal{V}_y $ can be utilized to represent class $y$,  we use $h = max()$ to calculate the final probability score of each class.
Then the probability score of each class $y$ can be calculated as:
\begin{equation}
\mathrm{Pr}(y \mid x ) 
= h (\mathrm{P}(v_y^1,x), \mathrm{P}(v_y^2,x), ..., \mathrm{P}(v_y^{k_y},x))
\end{equation}
where for $(x, v_y^i) $ that satisfies $(x, v_y^i) \in \tilde{\mathcal{D}}_{\text{train}} $ and $ v_y^i \in \mathcal{V}_y, (i = 1,2,...,k_y) $, $\mathrm{P}(v_y^i,x) $ is denoted as the conditional probability of filling the position of $\texttt{[mask]}$ with $v_y^i$:
\begin{equation}
\mathrm{P}(v_y^i,x) = \mathrm{Pr}(\texttt{Po([mask])} = v_y^i \mid x )
\end{equation}
After we obtain the probability score over each class, the final predicted class $\hat{y}$ is calculated as:
\begin{equation}
\hat{y}=\operatorname{argmax}_{y \in \mathcal{Y}} \mathrm{Pr}(y \mid x ) 
\end{equation}

%% file: experiments.tex
\section{Experiments}\label{sec:exp}

In this section, we present the experiments to evaluate the effectiveness of the proposed {\m}. Specifically, we aim to answer the following research questions:
\begin{itemize}
    \item \textbf{RQ1} Can {\m} improve the performance of few-shot prompt-based tuning?
    \item \textbf{RQ2} Can the proposed Label Augmentation strategy help  the target label prediction?
    \item \textbf{RQ3} Can {\m} make the prompt-based tuning method more stable?
\end{itemize}

\subsection{Experimental Settings}
\paragraph{Datasets.} We evaluate the proposed framework on few-shot text classification datasets from the widely-used NLU benchmark GLUE~\cite{wang2018glue} including SST-2~\cite{socher2013recursive}, CoLA~\cite{warstadt2019neural} and other common datasets including MR~\cite{pang2005seeing}, CR~\cite{hu2004mining}, Subj~\cite{pang2004sentimental},  MPQA~\cite{wiebe2005annotating}, SST-5~\cite{socher2013recursive}. These datasets cover different tasks such as sentiment analysis, topic classification, and opinion classification from various domains including movie reviews, news pieces, etc. The statistics of the datasets are shown in Table~\ref{tab:data_appendix} in the Appendix.

\paragraph{Baselines.} 
We compare the proposed approach with various representative methods:

\begin{itemize}
    \item \textbf{Majority}: The label is predicted by taking the majority  class in the training set. We run this baseline on the full-data setting.
    \item \textbf{Fine-Tuning}: The prediction is based on the pre-trained language model that is fine-tuned with the specific training data. We run this baseline in the full-data and few-shot settings.
    \item  \textbf{$\text{GPT-3}$}~\cite{brown2020language}: GPT-3 in-context tuning with a few samples. We pack the training samples into the input together and directly conduct inference.
    
     \item \textbf{EFL}~\cite{wang2021entailment}: An entailment-based prompt tuning framework. For a fair comparison, we do not pretrain the language model on MNLI task but directly tune the language model as the prompt tuning paradigm.
    \item \textbf{LM-BFF}~\cite{gao2021making}: 
    A prompt tuning model that automatically searches for demonstrations, templates, and label words. Note that LM-BFF utilizes a one-to-one verbalizer for label word selection.
    \item \textbf{Prompt Tuning}: The standard Prompt-based Tuning augmented by a simple template or template-free.
\end{itemize}

\begin{table*}[t]
\centering
\small
\begin{tabular}{lcccccccccc}
\toprule
 \textbf{Method} & \textbf{SST-2}  & \textbf{MR}  & \textbf{CR} & \textbf{Subj} & \textbf{CoLA} & \textbf{MPQA} &  \textbf{SST-5} \\
 
  & (Acc)  & (Acc)   & (Acc)  & (Acc)  & (Acc)  & (Acc)  & (Acc)  \\
  
\midrule
Majority (full)   & 50.9   & 50.0 & 50.0  & 50.0 & 69.1 & 50.0 & 23.1\\
Fine-Tuning (full)  & 95.0     & 90.8    & 89.4  & 97.0  & 86.2 & 89.4 & 58.7 \\

\midrule
\multicolumn{4}{l}{\emph{Few-shot scenario with K=8}} \\
\midrule

Fine-Tuning & 60.5 (3.1) & 60.3 (7.5)  & 61.9 (5.1)     & 78.3 (8.2)  & 51.1 (8.4) & 59.0 (3.4) & 31.5 (7.5) \\

$\text{GPT-3}$~\cite{brown2020language}       & 82.9 (3.4)  & 81.2 (2.5)  & 86.8 (1.5)     & 53.2 (1.5) &  52.1 (6.2) & 62.9 (3.5)  & 31.5 (4.3)  \\

$\text{EFL}$~\cite{wang2021entailment}    & 67.5 (8.5)  & 69.8 (7.5)  & 75.3 (4.8)     & 78.9 (7.8) & 54.3 (8.9) & 68.4 (5.7)  & 35.2 (6.3)  \\

$\text{LM-BFF}$~\cite{gao2021making}          & 89.1 (4.1)  & 83.6 (3.4)  & 87.8 (4.3)     & 81.6 (6.1) & 53.5 (4.5) & 73.9 (8.9)  & 41.2 (3.1)  \\

\midrule
$\text{Prompt Tuning}^{\ddagger}$ & 85.5 (5.2)  &  83.0 (3.7)                &      86.5 (3.0)   &   81.8 (5.6) & 50.5 (10.3) & 71.5 (9.8) & 37.5 (5.5)  \\
$\text{PT}$ + ${\m} (\text{m.})^{\ddagger}$   &    87.3 (4.4)   &  82.5 (1.4)  &   88.1 (2.7)           &     81.3 (4.9)  & 51.2 (7.5)  & 72.9 (9.1) & 39.4 (4.3)     \\      
$\text{PT}$ + ${\m} (\text{au.})^{\ddagger}$    &    87.6 (4.1)       &   83.1 (3.1)        &   87.8 (1.2)                 &    83.4   (2.5)    & 52.8 (8.1) & 74.5 (7.8)  & 41.8 (3.9)               \\    

\midrule
$\text{Prompt Tuning}^{\dagger}$ & 85.8 (5.8)  &   79.3 (8.2)                &      86.1 (8.0)  &  81.2 (5.7)  & 52.7 (6.6) & 75.1 (13.7) & 38.4 (4.7) \\
 $\text{PT}$ + ${\m} (\text{m.})^{\dagger}$   &    88.9 (3.9)                   &        \textbf{83.8 (1.9)}               &      84.9 (5.7)           &        82.4 (9.9)   & 51.3 (15.5) & 78.1 (8.9) & 42.7 (7.1)  \\      

\rowcolor{gray!20}$\text{PT}$ + ${\m} (\text{au.})^{\dagger}$    &  \textbf{89.5 (2.9)}  & 83.7 (2.6) &	\textbf{88.3 (4.1)}	&	\textbf{86.8 (3.1)}   & \textbf{55.9 (7.1)} & \textbf{78.4 (9.2)}   & \textbf{43.3 (1.6)}      \\    

\bottomrule
\end{tabular}

\caption{The main results using RoBERTa-large on representative NLU tasks. All the results are evaluated on full test sets and averaged over 5 runs. $K=8$: 8 samples per class for all the experiments; $^\dagger$: template augmented; $^\ddagger$: template-free; (m.): manual label augmentation; (au.): automatic label augmentation; PT: Prompt Tuning. 
}
\label{tab:results}

\end{table*}

\paragraph{Evaluation setting.}  Evaluation is critical in few-shot scenarios because small changes of the training set can result in a large variance in the performance on the test set. Following the few-shot setting in \cite{DBLP:conf/nips/PerezKC21, zhang2020revisiting, gu2021ppt, gao2021making}, we randomly select $K$-shot samples from original datasets for each class to construct the training set $\mathcal{D}_{\text{train}}$ and select another $K$-shot samples to construct the development set $\mathcal{D}_{\text{val}}$. For enhancing the stability of evaluation, we utilize the whole test set of original datasets as our test set $\mathcal{D}_{\text{test}}$ and change the random seed of sampling $\mathcal{D}_{\text{train}} $ and $\mathcal{D}_{\text{val}}$ for five times. 
We select RoBERTa-large as our backbone model to make a fair comparison with the baseline LM-BFF.

\subsection{Experimental Results}
In this subsection, we present our main results and address the aforementioned research questions pertaining to our {\m} approach.

In addition to comparing with baselines such as Majority, normal fine tuning, and prompt-based methods GPT-3, EFL, LM-BFF, we conduct more experiments to verify the effectiveness of our proposed method {\m} as a plug-in module. Because different template choices can result in a large variance of performance~\cite{gao2021making}, we design two groups of experiments, namely \textit{template-free} and \textit{template-augmented}, to investigate whether or not our method can improve over standard prompt-based tuning method regardless of the template design. For the template-augmented group of experiments, we manually choose ``\texttt{It is [MASK]}" as the template, following \cite{wang2021entailment}. For the template-free group of experiments, we only append ``\texttt{[MASK]}" in the input. We report the results of {\m} in Table~\ref{tab:results} when the size of data augmentation is $\times 3$ (i.e., $k_y=3$). We also consider two scenarios where the label words are derived \textit{manually} or  with our \textit{automatic} label augmentation mechanism.
We choose 8 samples ($K=8$) per class as the few-shot setting of our main experiments. For a fair comparison, we choose the same random seeds of training set sampling as LM-BFF. We train for 10 epochs on each dataset following \cite{wang2021entailment}. We report the average performance and standard deviation of our results over five runs of sampling on each dataset. The main results can be seen in Table~\ref{tab:results}.

\begin{table*}[ht]
\centering
\small

\begin{tabular}{lcccccccccc}
\toprule
  \textbf{Method}  &  \textbf{SST-2}  &   \textbf{MR}  &   \textbf{CR} &   \textbf{Subj}   &   \textbf{SST-5}  \\
  
   & (Acc)  & (Acc)   & (Acc)  & (Acc)    & (Acc)    \\
  
\midrule

PT   & 85.8 (5.8)  &   79.3 (8.2)                &      86.1 (8.0)  &  81.2 (5.7)  & 38.4 (4.7) \\

PT with Conventional DA  & 89.2 (1.3)  &   80.3 (3.1)                &      86.5 (4.5)  &  82.3 (8.0)  &  39.1 (4.5)    \\

PT with {\m} &  89.5 (2.9)  & 83.7 (2.6) &	88.3 (4.1)	&	86.8 (3.1)   & 43.3 (1.6)  \\

\rowcolor{gray!20}PT with {\m} \& Conventional DA  & \textbf{89.7 (1.6)}  &  \textbf{84.8 (1.5)}    & \textbf{89.2 (1.3)}    & \textbf{87.0 (3.1)}   & \textbf{43.6 (1.1)}    \\

\bottomrule
\end{tabular}

\caption{The main results of evaluating Prompt Tuning (PT) with {\m} and conventional DA method on NLU tasks. All the results are evaluated on full dev sets and averaged across 5 different training sets. $K=8$: 8 samples per class for the experiments. Conventional DA refers to \textit{synonym substitution}.}
\label{tab:table3}

\end{table*}

\paragraph{Performance analysis}
We analyze the performance from three perspectives to answer the aforementioned research questions. 

To answer \textbf{RQ1}, we compare the proposed method with existing baselines. First, in general, we can observe that the standard prompt-based tuning method with {\m} consistently performs better than or is comparable with  baselines such as GPT-3, EFL, LM-BFF, and normal fine tuning (results of ``$\text{PT}$ + ${\m} (\text{au.})^{\dagger}$'' in Table~\ref{tab:results}). Compared with LM-BFF, standard prompt-based tuning with {\m} performs better on all the datasets. For example, our method achieves a more than 6\% gain over LM-BFF on Subj and MPQA. Compared with normal fine tuning, our method achieves superior performance by a large margin. For example, our method obtains a 47.9\% improvement over normal fine tuning on SST-2.

Second, we can see that {\m} can improve over
standard prompt-based tuning method regardless of template design (results of ``$\text{PT}$ + ${\m} (\text{au.})^{\dagger}$'' and ``$\text{PT}$ + ${\m} (\text{au.})^{\ddagger}$'' in Table~\ref{tab:results}). Compared with standard prompt tuning, {\m} can achieve better performance over the seven datasets regardless of 
being template-free or template-augmented, which suggests that {\m} has no relation with template design and can be used as a plug-in module for improving the performance of prompt tuning.

Third, {\m} generally improves over the standard prompt tuning method regardless of automatic label word selection or manual label word selection (results of ``$\text{PT}$ + ${\m} (\text{au.})^{\dagger}$'' and ``$\text{PT}$ + ${\m} (\text{m.})^{\dagger}$'' in Table~\ref{tab:results}). Compared with standard prompt tuning, prompt tuning with automatic label word selection achieves improvements over all the datasets. For prompt tuning with manual label word selection, it also has performance gain over most datasets including SST-2, MR, Subj, MPQA, and SST-5.

To answer \textbf{RQ2}, we perform an ablation study of {\m}, and compare the results of ``$\text{PT}$ + ${\m} (\text{au.})^{\dagger}$'' and ``$\text{PT}$ + ${\m} (\text{m.})^{\dagger}$'' in Table~\ref{tab:results}. We can see that regardless of template design, our proposed automatically searched label words generally perform better than manually searched label words. For example, ``$\text{PT}$ + ${\m} (\text{au.})^{\dagger}$'' achieves a 5.3\% improvement over  ``$\text{PT}$ + ${\m} (\text{m.})^{\dagger}$'' on Subj dataset.

We analyze the reason from three perspectives. \textit{First}, we hypothesize that human bias may hinder selecting optimal label words and our proposed automatic method relies on the language model itself and can minimize human bias. \textit{Second}, it may be easier for humans to select similar words  as label words for sentiment-related datasets with the label name ``\texttt{positive, negative}'', but it is hard to select semantically similar words as label words for tasks in other domains. For example, it is hard to manually identify semantically similar words as label words for the Subj dataset with the label name ``\texttt{subjective, objective}'', which illustrates the necessity of our proposed automatic method for searching label words.  \textit{Third}, our proposed Label Augmentation method can search for different label words on different training data, but it is hard for the manual label word selection method to adapt to different specific datasets.

To answer \textbf{RQ3}, we analyze the stability of performance of {\m}. In general, we observe that {\m} reduces the variance of prompt-tuning. (Standard deviation of ``$\text{PT}$ + ${\m} (\text{au.})^{\dagger}$'' and ``$\text{Prompt Tuning}^{\dagger}$'' in Table~\ref{tab:results}). 
The uncertainty of prompt-based tuning methods mainly comes from different distributions of the small training sets, different designs of the template, and various selections of label words for each class.
Compared with the standard prompt-based tuning method, {\m} can improve the stability of tuning on most of the datasets. For example, the standard deviation of prediction over five runs for ``$\text{PT}$ + ${\m} (\text{au.})^{\dagger}$'' has decreased 50.0\% on SST-2 and 66.0\% on SST-5 respectively compared with ``$\text{Prompt Tuning}^{\dagger}$''.

\begin{table*}[tbp!]
\centering 
\renewcommand\arraystretch{1}
\resizebox{\linewidth}{!}{
\begin{tabular}{lll}
\toprule
\multirow{3}{*}{\textbf{SST-2}} &label name & \texttt{positive | negative}  \\
\noalign{\vskip 0.3ex}
\cline{2-3}
\noalign{\vskip 0.3ex}
&label words (m.)  & \texttt{positive, great, good | negative terrible bad} \\
\noalign{\vskip 0.3ex}
\cline{2-3}
\noalign{\vskip 0.3ex}
&label words (au.) & \texttt{wonderful brilliant fantastic | terrible done disappointing} \\

\midrule
\multirow{3}{*}{\textbf{Subj}} & label name & \texttt{objective | subjective}\\
\noalign{\vskip 0.3ex}
\cline{2-3} 
\noalign{\vskip 0.3ex}
&label words (m.)  & \texttt{good neutral fair | bad emotional personal} \\

\noalign{\vskip 0.3ex}
\cline{2-3}
\noalign{\vskip 0.3ex}
&label words (au.) &  \texttt{disturbing terrifying key | bad not nonsense} \\

\midrule
\multirow{5}{*}{\textbf{SST-5}} & label name & \texttt{very positive | positive | neutral | negative | very negative}\\
\noalign{\vskip 0.3ex}
\cline{2-3}  
\noalign{\vskip 0.3ex}
& \multirow{2}{*}{label words (m.)} &  \texttt{great perfect excellent | good, pretty, wonderful |} \\
&&  \texttt{neutral normal fine | bad worse not | terrible awful ridiculous} \\

\noalign{\vskip 0.3ex}
\cline{2-3}
\noalign{\vskip 0.3ex}
& \multirow{2}{*}{label words (au.)} &  \texttt{great brilliant fantastic | extraordinary remarkable fascinating |} \\
&&  \texttt{enough terrible funny | awful bad worse | boring done unnecessary} \\

\bottomrule
\end{tabular}
}

\caption{An illustration of the label words searched automatically or manually on SST-2, Subj and SST-5 datasets.}
\label{tab:words}

\end{table*}

\subsection{{\m} vs. Conventional DA}
\label{sec:Combination_with_Conventional_DA}

Although conventional data augmentation methods are mostly effective when training data is limited~\cite{chen2021empirical},  previous works have demonstrated that they can only bring marginal improvement for the prompt tuning paradigm~\cite{zhou2021flipda}.
Thus, in this subsection,  we conduct further experiments to compare {\m} and conventional DA. In addition, we also explore the performance of their combination.

We follow the same setting as the standard prompt-based tuning paradigm with template in the main experiments and utilize {\m} to enlarge the training set by  $\times 3$. With regards to conventional DA, we select the representative augmentation method \textit{synonym substitution}  from nlpaug toolkit~\cite{ma2019nlpaug} and also enlarge the training set by $\times 3$. The results are shown as Table~\ref{tab:table3}.

\paragraph{Comparison with Conventional DA}
We compare the performance of {\m} and Conventional DA on five datasets and can observe that {\m} consistently outperforms Conventional DA  by a large margin. For example, {\m} has a 10.74\% performance gain over synonym substitution on SST-5. The results further demonstrate the effectiveness of incorporating label semantics in augmentation for prompt tuning.

\paragraph{Combination with Conventional DA}
We can observe that the combination of {\m} and Conventional DA method has a consistent improvement over only using {\m} or Conventional DA.  Conventional DA methods such as synonym substitution mostly focus on exploiting the semantic information of the training instance itself. Our method proposes to utilize label semantic information to guide data augmentation and does not change instances. Our proposed {\m} conducts the augmentation from a different perspective compared with conventional DA methods.  Therefore, {\m} can be regarded as \textit{orthogonal} to conventional augmentation strategies and complement them to further improve the performance.

\subsection{Analysis of Label Word Selection}

In this subsection, without loss of generosity, we take the datasets SST-2, Subj, and SST-5 for example to analyze the quality of Label Augmentation (the label word results are shown in Table~\ref{tab:words} and the complete results over five runs on SST-2, CR, MR, Subj,
CoLA, MPQA, and SST-5 datasets are shown in Appendix Table~\ref{tab:label_words}). \textit{The goal of label augmentation is to find semantically similar words to enrich the label space.} With regards to the manual way, we find the synonyms of the label name from a dictionary as the label words and ensure these words are in the vocabulary. And we select the same label words for different seeds. As for our proposed automatic method, we only rely on the training set and language model (e.g., RoBERTa-large) to find semantically similar words from vocabulary.

Table~\ref{tab:words} shows the  label words automatically or manually searched on datasets SST-2, Subj, and SST-5 respectively. For sentiment-related datasets such as SST-2 with the label name \{$\texttt{positive/negative}$\}, the label words automatically searched are literally similar to the manually selected label words, which probably means the way language models (e.g., RoBERTa-large) reasons about what are similar words is close to the human way \textit{in sentiment domain}. Nonetheless, for other datasets such as Subj with the label name \{$\texttt{objective/subjective}$\}, it is interesting to observe that the label words automatically searched are not literally similar to label name or manually selected label words, which may infer that  the way language models reason about word semantic similarity in a different way \textit{in other domains}. It is important to  study \textit{how to define word similarity in label semantic space} in the future. For datasets such as SST-5, we can see that it is much harder to select appropriate label words when the number of classes is larger, which also verifies the importance of automatic label word selection.

\subsection{Impact Factors of {\m}}

\begin{figure}[tp!]
\centering
\hspace{-1.5em}
\subfigure[\# size of augmentation]{
  {\includegraphics[width=0.24\textwidth]{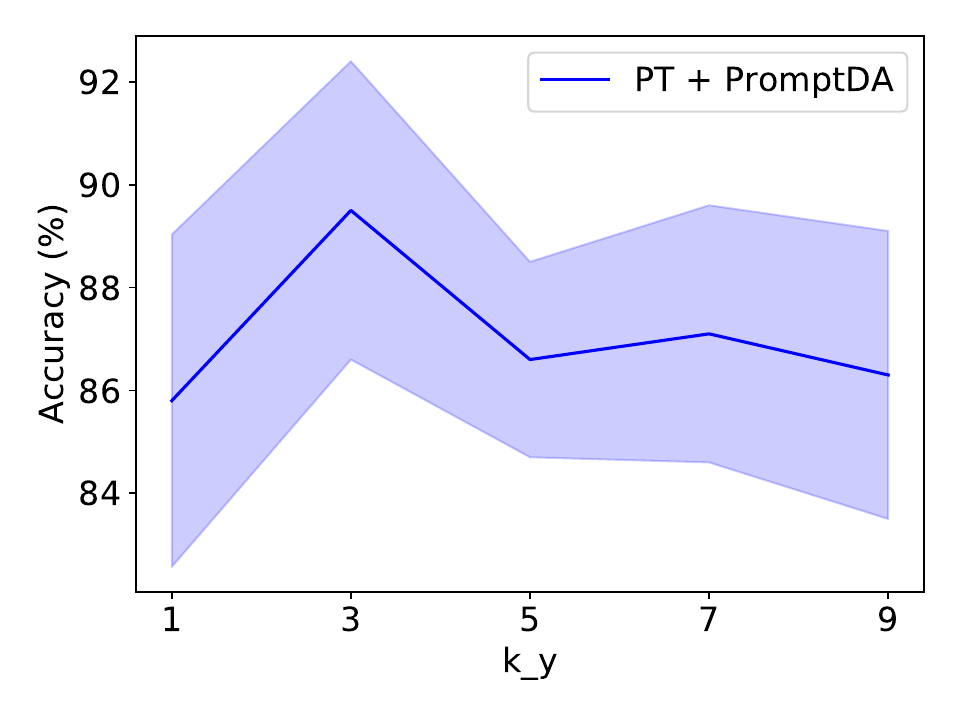}}
}
\hspace{-0.25cm}
\subfigure[\# samples per class]{
  {\includegraphics[width=0.24\textwidth]{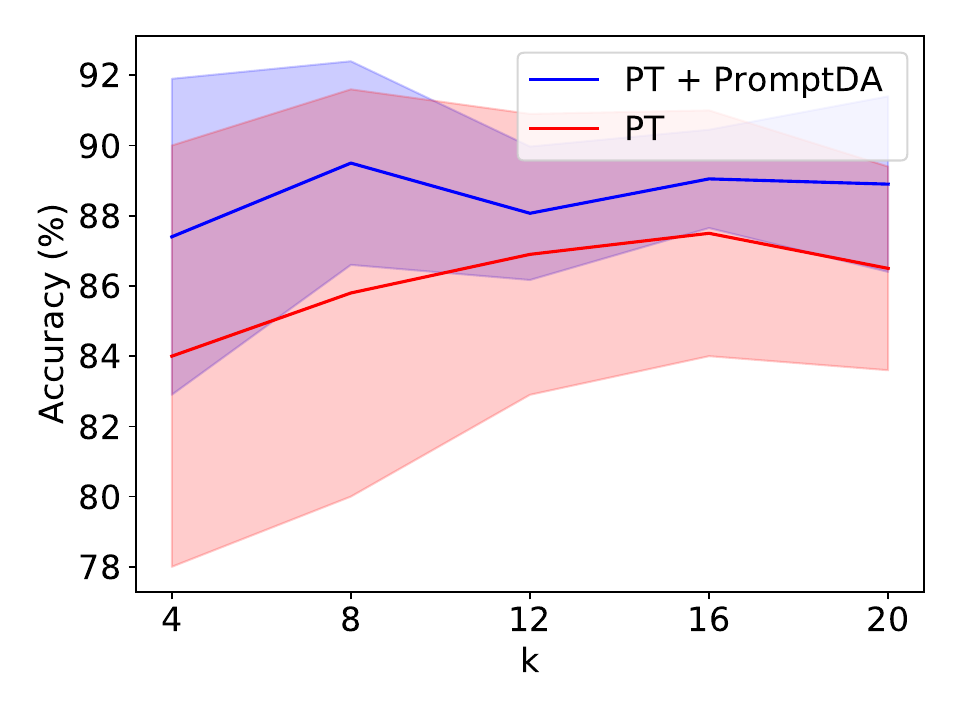}}
}
\caption{The impact analysis of the size of label words and training samples per class on SST-2 dataset.}\label{fig:size}
\label{Assessment}
\end{figure}

In this subsection, we analyze the impact factors of {\m} from two perspectives including the size of data augmentation and the size of training set. 
The results are shown in Figure~\ref{Assessment}.

\paragraph{The size of data augmentation}
We choose to study the effect of the size of {\m} on template-augmented prompt-based tuning on SST-2 dataset. There are 8 samples per class in the training set. The results over five runs for 10 epochs are presented in Figure~\ref{Assessment} (a). 
We can observe that {\m} can generally improve over prompt-based tuning regardless of the size of augmentation. However, larger augmentation may result in a more unstable final prediction. We analyze the reason from two perspectives. \textit{First}, larger data augmentation may contain more label noise. Since we utilize a one-to-multiple verbalizer to guide data augmentation, the size of data augmentation is equal to the number of label words per class, which may cause more noisy label words. Unsuitable label word selections may  worsen the performance and increase the variance of the final prediction. \textit{Second}, more label words per class may cause the model harder to converge on small training sets. When training for the same epochs, prompt tuning with more label words per class may perform more unstable.

\paragraph{The size of the training set}
We study the effect of the size of the training set on template-augmented prompt-based tuning with and without {\m}. The size of data augmentation is $\times 3$. The results over five runs for 10 epochs are presented in Figure~\ref{Assessment} (b). We have several observations from the results. \textit{First}, our method {\m} consistently improves over standard prompt-based tuning regardless of the size of training sets, which demonstrates the effectiveness of {\m} regardless of the size of the training set in the few-shot scenario.  \textit{Second}, our proposed method generally decreases the variance of prompt-based tuning. This further shows that our method can stabilize the training process of prompt tuning.  \textit{Third}, the improvement space of {\m} over prompt-based tuning decreases as the number of samples per class increases. Thus, our method tends to be more effective when the training samples are very limited.

%% file: appendix.tex
\section{Appendix}

\begin{table*}[h]
\centering 
\small
\begin{tabular}{lcccccc}
\toprule
Dataset &  \# Classes & \# Length &  \# Train  &  \# Test & Type & Labels \\
\midrule

SST-2 & 2 & 19 & 6,920  & 872 & sentiment & positive, negative \\
MR & 2 & 20 & 8,662 & 2,000 &  sentiment & positive, negative \\
CR & 2 & 19 & 1,775 & 2,000 & sentiment & positive, negative \\
Subj & 2 & 23 & 8,000 & 2,000 & subjectivity & subjective, objective \\
CoLA & 2 & 8 & 8,551 & 1,042 & acceptability & grammatical, not\_grammatical \\
MPQA & 2 & 3 & 8,606 & 2,000 & opinion polarity & positive, negative \\
SST-5 & 5 & 18 & 8,544 & 2,210 & sentiment & v. pos., positive, neutral, negative, v. neg. \\

\bottomrule
\end{tabular}
\caption{Statistics of the datasets. }
\label{tab:data_appendix}
\end{table*}

\subsection{Implementation Details}

We implemented our model and all baselines with PyTorch and run each experiment on a single NVIDIA GeForce RTX 3090 GPU. The hyperparameters are the same for  all methods based on RoBERTa-large (the learning rate is 3e-6, the batch size is 4, and the number of training epochs is 10). Following \cite{gao2021making}, we select the random seeds for sampling the training set and validation set as $\{13, 21, 42, 87, 100\}$. 

\subsection{Dataset Details}
\label{sec:Dataset_Details}

In general, we follow the experiment setting of \cite{gao2021making}.
For datasets from GLUE~\cite{wang2018glue}  including SST-2~\cite{socher2013recursive} and CoLA~\cite{warstadt2019neural}, we use the original development sets for testing. For datasets requiring cross-validation evaluation like MR~\cite{pang2005seeing}, CR~\cite{hu2004mining}, MPQA~\cite{wiebe2005annotating} and Subj~\cite{pang2004sentimental}, we randomly sample 2,000 instances as the testing set and remove them from the training set. For the dataset SST-5~\cite{socher2013recursive}, we use the official test sets. The dataset statistics are shown in Table~\ref{tab:data_appendix}.

\subsection{Comparison of RoBERTa vs BERT}

We conduct experiments to investigate the impact of the backbone model.
Table \ref{tab:BERT_large} shows the results of using BERT-large(uncased) and RoBERTa-large. The experiment setting is the same as the main experiments. We can observe that our proposed {\m} improves the performance of prompt tuning regardless of the backbone model.

\begin{table}[h]
\centering 
\small
\begin{tabular}{lcccccc}
\toprule

\textbf{BERT-large} &  \textbf{SST-2} & \textbf{Subj} &  \textbf{SST-5}  \\
\midrule

PT & 82.3 (4.6) & 80.3 (6.2) & 34.5 (3.8)\\

\rowcolor{gray!20}PT + {\m} & 87.1 (3.1) & 82.9 (3.3) & 37.5 (2.8) \\

\midrule

\textbf{RoBERTa-large} &  \textbf{SST-2}  & \textbf{Subj} &  \textbf{SST-5}  \\
\midrule

PT  & 85.8 (5.8)  & 81.2 (5.7) & 38.4 (4.7)\\

\rowcolor{gray!20}PT + {\m} & \textbf{89.5 (2.9)}  & \textbf{86.8 (3.1)} & \textbf{43.3 (1.6)} \\

\bottomrule
\end{tabular}
\caption{A comparison of RoBERTa-large vs BERT-large on template-augmented prompt tuning.}
\label{tab:BERT_large}
\end{table}

\subsection{The Verbalizer and Template Design}
For each random seed of $\{13, 21, 42, 87, 100\}$, we can construct different training sets and validation sets. Thus, the verbalizers searched automatically  for each run are different, which are shown as ``label words (au.)'' in Table~\ref{tab:label_words}. The verbalizers manually designed for each run are the same, which are shown as ``label words (m.)'' in Table~\ref{tab:label_words}. We follow previous works~\cite{gao2021making, wang2021entailment} and design a simple template  ``\texttt{It is [MASK]}'' for each input.

\begin{table*}[tbp!]
\footnotesize
\centering 
\resizebox{\linewidth}{!}{
\begin{tabular}{lll}
\toprule
\multirow{8}{*}{\textbf{SST-2}} &label name & \texttt{positive | negative}  \\

\cline{2-3}
&label word (s.)  & \texttt{positive | negative} \\

\cline{2-3}
&label words (m.)  & \texttt{good perfect fantastic | terrible awful hilarious} \\

\cline{2-3}
& \multirow{5}{*}{label words (au.)} & \texttt{brilliant amazing wonderful |  not awful terrible} \\

\cline{3-3}
&  & \texttt{great perfect brilliant | terrible disappointing bad} \\

\cline{3-3}
&  & \texttt{beautiful perfect fantastic | terrible awful hilarious} \\

\cline{3-3}
&  & \texttt{fantastic excellent beautiful | terrible awful worse} \\

\cline{3-3}
&  & \texttt{wonderful, brilliant, fantastic | terrible done disappointing} \\

\midrule

\multirow{8}{*}{\textbf{MR}} &label name & \texttt{positive | negative}  \\

\cline{2-3}
&label word (s.)  & \texttt{positive | negative} \\

\cline{2-3}
&label words (m.)  & \texttt{positive, great, good | negative, terrible, bad} \\

\cline{2-3}
& \multirow{5}{*}{label words (au.)} & \texttt{refreshing good beautiful | not terrible disappointing} \\

\cline{3-3}
&  & \texttt{beautiful perfect fantastic | awful disappointing horrible} \\

\cline{3-3}
&  & \texttt{fantastic wonderful beautiful | terrible awful funny} \\

\cline{3-3}
&  & \texttt{fantastic incredible unforgettable | terrible funny bad} \\

\cline{3-3}
&  & \texttt{excellent refreshing amazing | terrible wrong bad} \\

\midrule

\multirow{8}{*}{\textbf{CR}} &label name & \texttt{positive | negative}  \\

\cline{2-3}
&label word (s.)  & \texttt{positive | negative} \\

\cline{2-3}
&label words (m.)  & \texttt{good perfect fantastic | terrible awful hilarious} \\

\cline{2-3}
& \multirow{5}{*}{label words (au.)} & \texttt{amazing fun cool | disappointing frustrating bad} \\

\cline{3-3}
&  & \texttt{excellent fun cheap | awful horrible terrible} \\

\cline{3-3}
&  & \texttt{free fun cool | bad painful useless} \\

\cline{3-3}
&  & \texttt{fantastic brilliant incredible | terrible inevitable useless} \\

\cline{3-3}
&  & \texttt{amazing great awesome | terrible awful horrible} \\

\midrule

\multirow{8}{*}{\textbf{Subj}} & label name & \texttt{objective | subjective}\\

\cline{2-3}
&label word (s.)  & \texttt{actual | individual} \\

\cline{2-3} 
&label words (m.)  & \texttt{good neutral fair | bad emotional personal} \\

\cline{2-3}
& \multirow{5}{*}{label words (au.)} & \texttt{epic life America | madness not wrong} \\

\cline{3-3}
&  & \texttt{life history significant | right that great} \\

\cline{3-3}
&  & \texttt{what real interesting | me good great} \\

\cline{3-3}
&  & \texttt{fiction interesting America | wonderful great brilliant} \\

\cline{3-3}
&  & \texttt{disturbing terrifying key | bad not nonsense} \\

\midrule

\multirow{8}{*}{\textbf{CoLA}} &label name & \texttt{grammatical | not\_grammatical}  \\

\cline{2-3}
&label word (s.)  & \texttt{good | bad} \\

\cline{2-3}
&label words (m.)  & \texttt{positive correct good | negative wrong bad} \\

\cline{2-3}
& \multirow{5}{*}{label words (au.)} & \texttt{it wrong correct | ridiculous not good} \\

\cline{3-3}
&  & \texttt{different sad interesting | complicated hilarious scary} \\

\cline{3-3}
&  & \texttt{wrong interesting important | insane sad crazy} \\

\cline{3-3}
&  & \texttt{all good important | bad new impossible} \\

\cline{3-3}
&  & \texttt{how amazing normal | true him me} \\

\midrule

\multirow{8}{*}{\textbf{MPQA}} &label name & \texttt{positive | negative}  \\

\cline{2-3}
&label word (s.)  & \texttt{good | bad} \\

\cline{2-3}
&label words (m.)  & \texttt{good perfect fantastic | terrible awful hilarious} \\

\cline{2-3}
& \multirow{5}{*}{label words (au.)} & \texttt{possible necessary adopted | wrong bad dark} \\

\cline{3-3}
&  & \texttt{obvious awesome fun | then difficult gone} \\

\cline{3-3}
&  & \texttt{right fun decided | reported unfair rejected} \\

\cline{3-3}
&  & \texttt{accepted good great | unavoidable awful bad} \\

\cline{3-3}
&  & \texttt{different good amazing | wrong bad funny} \\

\midrule

\multirow{20}{*}{\textbf{SST-5}} & label name & \texttt{very positive | positive | neutral | negative | very negative}\\

\cline{2-3}
&label word (s.)  & \texttt{extraordinary | great | enough | boring | awful} \\

\cline{2-3} 
& \multirow{3}{*}{label words (m.)} &  \texttt{great perfect excellent | good pretty wonderful |} \\
&&  \texttt{neutral normal fine | bad worse not |} \\
&&  \texttt{terrible awful ridiculous} \\

\cline{2-3}
& \multirow{15}{*}{label words (au.)} &  \texttt{good excellent unforgettable | hilarious inevitable funny |} \\
&&  \texttt{different time interesting | predictable bad over |} 
\\
&&  \texttt{dreadful boring horrible} 
\\

\cline{3-3}
&  &  \texttt{magnificent unforgettable fantastic | refreshing remarkable sublime |} \\
&&  \texttt{disappointing bad hilarious | neither predictable inevitable |} 
\\
&&  \texttt{depressing pathetic unnecessary} 
\\

\cline{3-3}
&  &  \texttt{wonderful fantastic incredible | terrifying refreshing interesting |} \\
&&  \texttt{hilarious done easy | better disappointing predictable |} 
\\
&&  \texttt{disgusting ridiculous horrible} 
\\

\cline{3-3}
&  &  \texttt{magnificent excellent too | stunning unexpected refreshing |} \\
&&  \texttt{simple done interesting | boring there worse |} 
\\
&&  \texttt{ridiculous sad weird} 
\\

\cline{3-3}

&  &  \texttt{great brilliant fantastic | extraordinary remarkable fascinating |} \\
&&  \texttt{enough terrible funny | awful bad worse |}  
\\
&&  \texttt{boring done unnecessary}  
\\

\bottomrule
\end{tabular}
}
\caption{The verbalizer design (single label word (s.) for normal prompt tuning, label words manually designed (m.) and automatically searched (au.) for prompt tuning with {\m}) over five runs on SST-2, CR, MR, Subj, CoLA, MPQA, SST-5 datasets.}
\label{tab:label_words}
\end{table*}